\documentclass[runningheads]{llncs}
\usepackage[T1]{fontenc}
\usepackage{graphicx}
\usepackage{amstext}
\usepackage{amsmath}
\usepackage{bm}
\usepackage{amssymb}
\usepackage{tabularx}
\usepackage{booktabs}
\usepackage{array}
\usepackage{enumitem}
\newcolumntype{C}{>{\centering\arraybackslash}X}
\usepackage{subcaption}

\usepackage[capitalize]{cleveref}
\crefname{section}{Sec.}{Secs.}
\Crefname{section}{Section}{Sections}
\Crefname{table}{Table}{Tables}
\crefname{table}{Tab.}{Tabs.}

\begin{document}
\title{Gaussian Alignment for Relative Camera Pose Estimation via Single-View Reconstruction}
\titlerunning{Gaussian Alignment for Relative Camera Pose Estimation}
%
\author{Yumin Li\orcidID{0009-0002-1069-4934} \and
Dylan Campbell\orcidID{0000-0002-4717-6850}}

\authorrunning{Y. Li and D. Campbell}
%
\institute{Australian National University, Canberra, Australia \email{\{yumin.li,dylan.campbell\}@anu.edu.au}}
\maketitle              
\begin{abstract}
Estimating metric relative camera pose from a pair of images is of great importance for 3D reconstruction and localisation.
However, conventional two-view pose estimation methods are not metric, with camera translation known only up to a scale, and struggle with wide baselines and textureless or reflective surfaces.
This paper introduces GARPS, a training-free framework that casts this problem as the direct alignment of two independently reconstructed 3D scenes.
GARPS leverages a metric monocular depth estimator and a Gaussian scene reconstructor to obtain a metric 3D Gaussian Mixture Model (GMM) for each image.
It then refines an initial pose from a feed-forward two-view pose estimator by optimising a differentiable GMM alignment objective.
This objective jointly considers geometric structure, view-independent colour, anisotropic covariance, and semantic feature consistency, and is robust to occlusions and texture-poor regions without requiring explicit 2D correspondences.
Extensive experiments on the Real\-Estate10K dataset demonstrate that GARPS outperforms both classical and state-of-the-art learning-based methods, including MASt3R.
These results highlight the potential of bridging single-view perception with multi-view geometry to achieve robust and metric relative pose estimation.

\end{abstract}

\section{Introduction}

Estimating the relative pose between two camera views---comprising 3D rotation and translation---is a fundamental problem in computer vision, underpinning applications such as robot navigation, augmented reality, and 3D reconstruction \cite{hartley2003multiple,szeliski2022computer}.
In challenging indoor environments, textureless walls, repeated structures, and varying illumination often lead to ambiguous correspondences, making reliable pose estimation particularly difficult.

Classical two-view methods estimate pose by matching local features under epipolar constraints \cite{SIFT}, typically by computing the essential matrix using algorithms such as the 5-point method \cite{Nister2004}, which requires known camera intrinsics.
However, these methods may fail in textureless or low-feature environments.
While multi-view Structure-from-Motion (SfM) pipelines like COLMAP \cite{COLMAP,COLMAP_MVS} provide more accurate reconstructions by jointly optimising multiple views, they require multiple images and significant computation, limiting their use in lightweight, pairwise scenarios.
Both approaches suffer from scale ambiguity.

The recent advent of single-view 3D reconstruction has created a new paradigm.
Methods like Flash3D \cite{Flash3D} can now generate a dense, explicit 3D scene representation from a single RGB image by leveraging powerful monocular depth estimators.
The speed and accuracy of single-view scene reconstructors makes an alternative approach more feasible: instead of matching ambiguous 2D pixels, we can directly align the reconstructed 3D models.

\begin{figure}[!t]
  \centering
  \includegraphics[width=\linewidth]{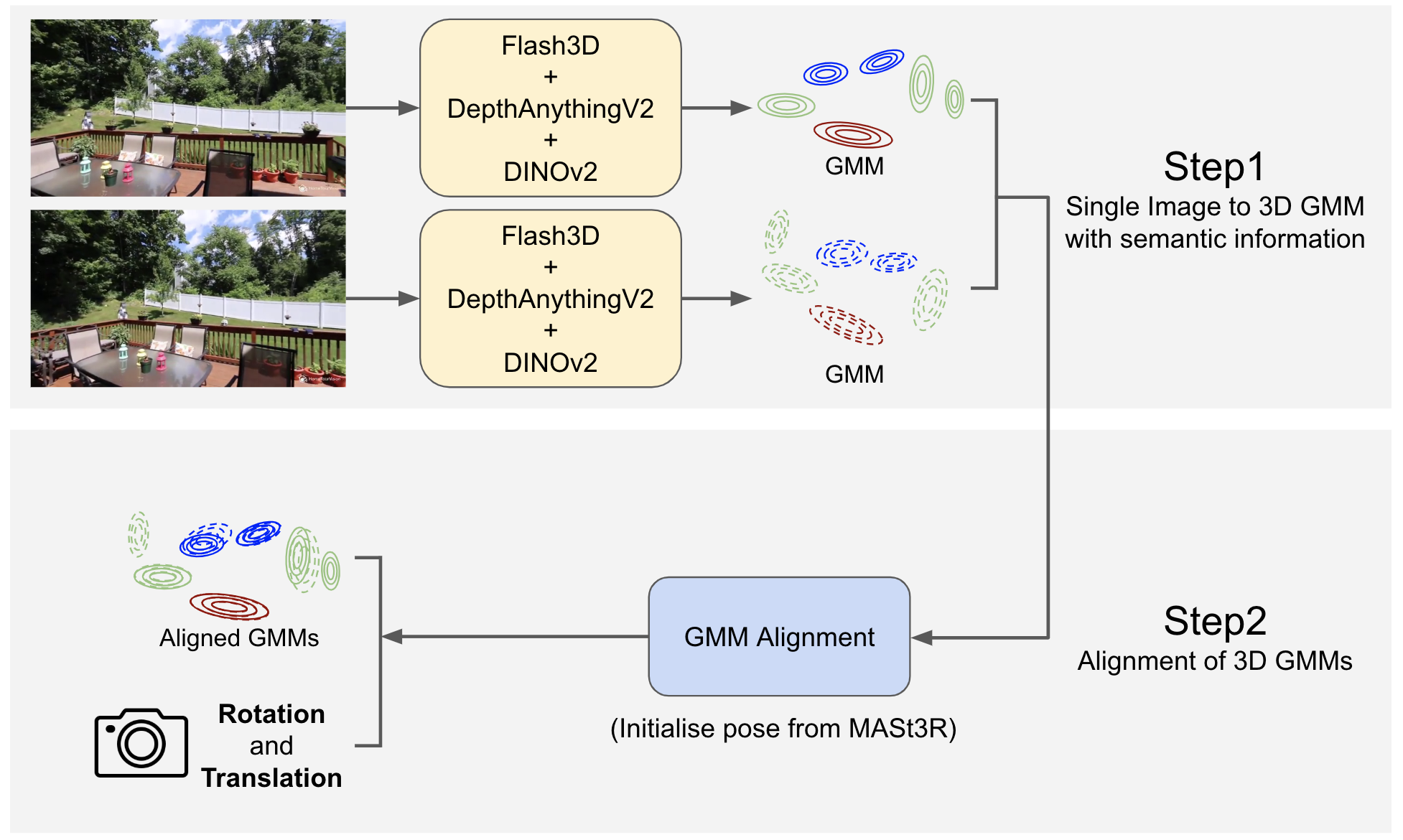}
  \caption{Overview of the proposed GARPS pipeline. Given two RGB images, we predict a 3D Gaussian mixture model (GMM) for each. The relative pose is then estimated by directly aligning these GMMs in 3D space using our novel optimisation objective.}
  \label{fig:main_process}
\end{figure}

Motivated by this, we introduce GARPS---Gaussian Alignment for Relative Pose via Single-view reconstruction---a novel method that casts relative pose estimation as an alignment problem between two 3D Gaussian Mixture Models (GMMs).
As shown in \cref{fig:main_process}, we first reconstruct each image into a metrically-scaled GMM.
We then directly optimise for the relative pose by maximising a differentiable objective function that jointly considers geometric structure, anisotropic covariance, view-independent colour, and semantic features.
Our approach bypasses the need for explicit keypoint matching and is training-free.

GARPS is inherently robust to outliers and local ambiguities compared to classical ICP \cite{ICP,ColoredICP} and, unlike learning-based methods such as DUSt3R \cite{DUSt3R}, it is interpretable and applicable to unseen domains without fine-tuning.
We make the following contributions:%
\begin{enumerate}[nosep]
\item GARPS, a novel method for relative pose estimation based on 3D GMM alignment from monocular reconstructions with metric scale; and
\item an alignment objective that combines 3D geometry, view-independent colour, anisotropic covariance and semantic features for robust pose optimisation.
\end{enumerate}
We evaluate our method on the RealEstate10K dataset \cite{RealEstate10K}, achieving state-of-the-art performance compared to both classical and learning-based baselines.

\section{Related Work}\label{sec:related}

Our work builds on recent progress in estimating relative camera pose and reconstructing 3D scenes from a single image.
In this section, we review related work in these two areas and explain how our method fits into this context.

\subsection{Two-View Relative Pose Estimation}

Estimating relative pose from two views is a long-standing problem in computer vision. We categorise prior work into three main streams.

Correspondence-based methods match 2D features, from classical descriptors \cite{SIFT} with robust estimators \cite{RANSAC}, to modern learned matchers \cite{SuperPoint,SuperGlue,LightGlue}.
These correspondences form the foundation for full Structure-from-Motion (SfM) pipelines like COLMAP \cite{COLMAP,COLMAP_MVS}, which leverage multi-view bundle adjustment to achieve globally consistent reconstructions.
However, these approaches require robust image features for matching and perform less well in areas that are texture-poor, have view-dependent appearance, or have local or global symmetries.

3D registration methods, in contrast, operate on point clouds.
The Iterative Closest Point (ICP) algorithm \cite{ICP} and its variants \cite{ColoredICP} are widely used but are prone to local minima.
Probabilistic approaches \cite{CPD,GMMReg} and globally-optimal solvers \cite{Go-ICP,GOGMA} offer more robustness.
However, they can be computationally expensive or rely on simplifying assumptions, such as GOGMA's \cite{GOGMA} use of isotropic covariances.

Learning-based methods directly regress pose or leverage learned scene representations.
Some infer dense 3D pointmaps from which pose is recovered \cite{DUSt3R,MASt3R}.
Others use structures like 3D Gaussians for pose estimation or refinement \cite{GS-CPR,Flare,LoopSplat}.
Despite their success, the performance of these methods is tied to large-scale supervision, and recovering metric scale often requires specific training datasets.

Our work, GARPS, contributes a new perspective to this field.
By framing pose estimation as the alignment of two independently generated 3D models, it combines the robustness of probabilistic 3D registration with the benefits of modern single-view reconstruction, avoiding direct 2D correspondence matching and end-to-end learning.

\subsection{Single-View 3D Reconstruction}

Our approach is enabled by recent progress in single-view 3D reconstruction, which itself builds upon monocular depth estimation.

Monocular depth estimation aims to infer pixel-wise depth from a single image.
Deep learning methods have largely superseded classical geometric approaches \cite{Shape-from-Shading,Texture-Gradient}.
State-of-the-art models like Depth Anything V2 \cite{DepthAnythingV2} leverage large-scale training with teacher--student frameworks to produce highly accurate and robust metric depth maps, showing strong generalisation capabilities.

These 2.5D depth maps can be upgraded to full 3D models by using them as a high-quality prior for 3D reconstruction.
While optimisation-based reconstruction methods like NeRF \cite{NeRF} and 3DGS \cite{3DGS} require multiple views, feed-forward methods like Splatter Image \cite{SplatterImage} and Flash3D \cite{Flash3D} can generate an explicit 3D Gaussian-based scene representation from a single image and its predicted depth map.
Unlike 2.5D depth maps, which only describe the parts of a scene visible from a single viewpoint, these 3D representations can also infer hidden or occluded areas---for example, Splatter Image can predict the back side of an object that is not visible in the input image.

In this work, we leverage this pipeline as the foundation for our method.
We enhance the Flash3D framework by integrating it with the metric Depth Anything V2 backbone.
This provides us with a high-quality, metrically-scaled Gaussian mixture model for each input image, which serves as the input to our alignment-based pose estimation algorithm.

\section{Single-view Reconstuction for Relative Pose Estimation}\label{sec:method}

In this section, we propose GARPS---Gaussian Alignment for Relative Pose via Single-view Reconstruction---a method that estimates the metric relative pose between two RGB images by aligning their single-view 3D reconstructions.
Each image is independently processed using Flash3D~\cite{Flash3D} to obtain a coloured 3D Gaussian Mixture Model (GMM), capturing both geometry, appearance and semantic information.
We estimate the relative pose by aligning these two GMMs via a differentiable $L_2$ distance-based objective function.

\subsection{Objective Formulation}

Given two images, we first reconstruct them into two sets of 3D Gaussian components, denoted as source GMM $\mathcal{G}_x$ and target GMM $\mathcal{G}_y$.
Each GMM is defined as $\mathcal{G} = \{ (\bm{\mu}_i, \bm{\Sigma}_i, \bm{c}_i, \phi_i) \}_{i=1}^{K}$, where $\bm{\mu}_i$ is the mean, $\bm{\Sigma}_i$ the spatial covariance, $\bm{c}_i$ the RGB colour, and $\phi_i$ the mixture weight.
Our goal is to find the relative pose $(\mathbf{R}^\star, \mathbf{t}^\star)$ with rotation $\mathbf{R}^\star \in \mathrm{SO}(3)$ and translation $\mathbf{t}^\star \in \mathbb{R}^3$
that best aligns mixtures $\mathcal{G}_x$ and $\mathcal{G}_y$.

We formulate this as minimising the $L_2$ distance between the two GMM probability density functions.
Let $T(\mathcal{G}_x, \mathbf{R}, \mathbf{t})$ denote the transformation of the source GMM, where each component is transformed as $\bm{\mu}_i' = \mathbf{R}\bm{\mu}_i + \mathbf{t}$ and $\bm{\Sigma}_i' = \mathbf{R}\bm{\Sigma}_i\mathbf{R}^\top$.
The squared $L_2$ distance is given by
\begin{align}
    d(\mathbf{R}, \mathbf{t})^2 &= \int_{\mathbb{R}^3} \left( p(\bm{p}|T(\mathcal{G}_x, \mathbf{R}, \mathbf{t})) - p(\bm{p}|\mathcal{G}_y) \right)^2 d\bm{p} \notag \\
    &= C - 2 \int_{\mathbb{R}^3} p(\bm{p}|T(\mathcal{G}_x)) \, p(\bm{p}|\mathcal{G}_y) \, d\bm{p} ,
\label{eq:l2_dist_conf}
\end{align}
where $C$ is a constant independent of the transformation.
Maximising the cross-term is equivalent to minimising the $L_2$ distance.
This cross-term has a convenient closed-form solution for GMMs, given by
\begin{align}
    \int p(\bm{p}|T(\mathcal{G}_x)) \, p(\bm{p}|\mathcal{G}_y) \, d\bm{p} = \sum_{i=1}^K \sum_{j=1}^K \phi_i \phi_j \varphi(\bm{0} \mid \bm{\mu}_i' - \bm{\mu}_j, \bm{\Sigma}_i' + \bm{\Sigma}_j),
\end{align}
where $\varphi(\bm{0} \mid \bm{\mu}, \bm{\Sigma})$ denotes the multivariate Gaussian probability density function with mean $\bm{\mu}$ and covariance $\bm{\Sigma}$, evaluated at $\bm{0}$.

To incorporate appearance, we augment each spatial Gaussian with a colour distribution, also modelled as a Gaussian.
The joint probability density, shown in \cref{eq:gmm_cross_conf}, allows us to extend the cross-term to include colour similarity in the same way as geometric similarity.
Building on this, we also incorporate semantic information by defining a semantic consistency weight $\omega_{ij}^{\text{sem}}$ for each Gaussian pair $(i, j)$, such that
\begin{equation}
\omega_{ij}^{\text{sem}} =
\begin{cases}
1 & \text{if } \ell_i = \ell_j \\
\lambda_\text{sem} & \text{otherwise,}
\end{cases}
\end{equation}
where $\ell$ denotes the semantic label, and $\lambda_\text{sem} \in [0, 1)$ is a discount factor applied to pairs with mismatched semantics.

The final cross-term in the objective function is weighted by both colour similarity and semantic consistency, promoting alignment between structures that are geometrically, visually, and semantically consistent.
It is given by
\begin{align}
\label{eq:gmm_cross_conf}
    \sum_{i=1}^K \sum_{j=1}^K \omega_{ij}^{\text{sem}} \phi_i \phi_j \underbrace{\varphi(\bm{0} \mid \bm{\mu}_i' - \bm{\mu}_j, \bm{\Sigma}_i' + \bm{\Sigma}_j)}_{\text{Geometric Term}} \underbrace{\varphi(\bm{0} \mid \bm{c}_i - \bm{c}_j, \bm{\Sigma}^c_i + \bm{\Sigma}^c_j).}_{\text{Colour Term}}
\end{align}

By expanding the Gaussian expressions and removing constant factors, we derive our final objective function to be maximised:
\begin{align}
f(\mathbf{R}, \mathbf{t}) = \sum_{i=1}^{K} \sum_{j=1}^{K} 
\frac{\omega_{ij}^{\text{sem}} \phi_i \phi_j}{|\bm{\Sigma}_i' + \bm{\Sigma}_j|^{1/2}} \,
\exp\left( -\frac{1}{2} D_{ij}^\text{geo} - \frac{1}{2} D_{ij}^\text{col} \right),
\label{eq:final_obj_conf}
\end{align}
where $D_{ij}^\text{geo} = (\bm{\mu}_i' - \bm{\mu}_j)^\top (\bm{\Sigma}_i' + \bm{\Sigma}_j)^{-1} (\bm{\mu}_i' - \bm{\mu}_j)$ is the squared Mahalanobis distance in 3D space, and $D_{ij}^\text{col} = (\bm{c}_i' - \bm{c}_j)^\top (\bm{\Sigma}^c_i + \bm{\Sigma}^c_j)^{-1} (\bm{c}_i' - \bm{c}_j)$ is the corresponding distance in RGB space.

\begin{figure}[t]
    \centering
    \includegraphics[width=0.8\linewidth]{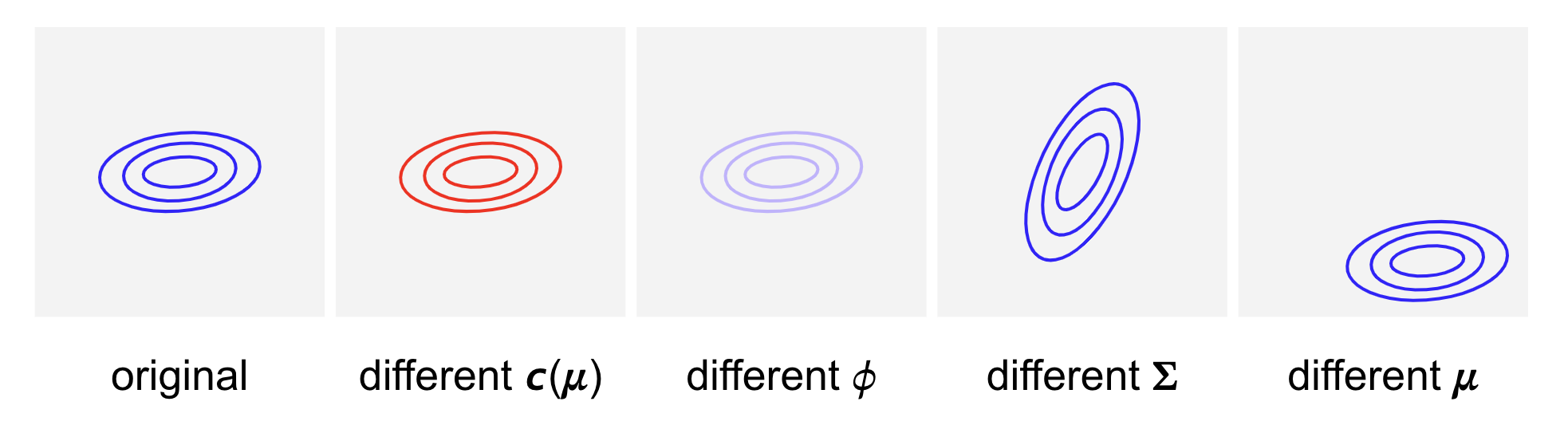}
    \caption{Visual effect of modifying Gaussian parameters: mean $\bm{\mu}$, covariance $\bm{\Sigma}$, opacity $\phi$, and colour $\bm{c}$.}
    \label{fig:gmm_visual}
\end{figure}

\paragraph{Component Formulation.}
The parameters for each Gaussian component are derived from a pre-trained Flash3D model:
the Gaussian means $\bm{\mu}_i$ and anisotropic spatial covariances $\bm{\Sigma}_i$ are copied directly from the Flash3D output, with the latter multiplied by a global scaling factor $\lambda_\text{cov}$ to modulate the sharpness of the geometric alignment term;
the colours $\bm{c}_i$ are taken from the view-independent zeroth-order spherical harmonics of Flash3D, lending robustness to reflective surfaces;
the colour covariances $\bm{\Sigma}^c$ are fixed diagonal hyperparameter matrices that control sensitivity to colour deviations; 
the mixture weights $\phi_i$ are derived from the Flash3D opacities after $L_1$ normalisation; and
the semantic labels $\ell$ are computed from a pre-trained DINOv2-based semantic segmentation network \cite{DINOv2} with 150 classes corresponding to the categories from the ADE20K dataset \cite{ADE20K,ADE20K_2}.
Specifically, we use the pretrained model (\texttt{dinov2\_vitg14\_ade20k\_m2f.pth}) released by the authors.
Note that Flash3D predicts two Gaussians per pixel, allowing it to model occluded surfaces.
We assign the same semantic label to both Gaussians, since usually these correspond to the front and back of the same object.
The components are visualised in \cref{fig:gmm_visual}.

\subsection{Optimisation Strategy}

To ensure computational efficiency, we operate on a sparse and salient subset of the original Gaussians.
This subset is formed by first discarding all components with an opacity value below a threshold $\tau_\text{opacity}$.
The remaining salient Gaussians are then uniformly subsampled with a stride of $s_\text{sub}$, which significantly reduces the number of components $K$ for optimisation.

Furthermore, directly evaluating the double summation in \cref{eq:final_obj_conf} is computationally prohibitive due to its $O(K^2)$ complexity.
To accelerate the process, we approximate the objective by considering only a sparse set of potential correspondences.
Specifically, for each source Gaussian, we find its
$k_\text{nn}$ nearest neighbours
in the target GMM based on the Euclidean distance between their means.
These candidate pairs are further pruned by discarding any for which this distance exceeds a threshold $\tau_\text{dist}$.

\begin{figure}[t]
    \centering
    \includegraphics[width=0.95\linewidth]{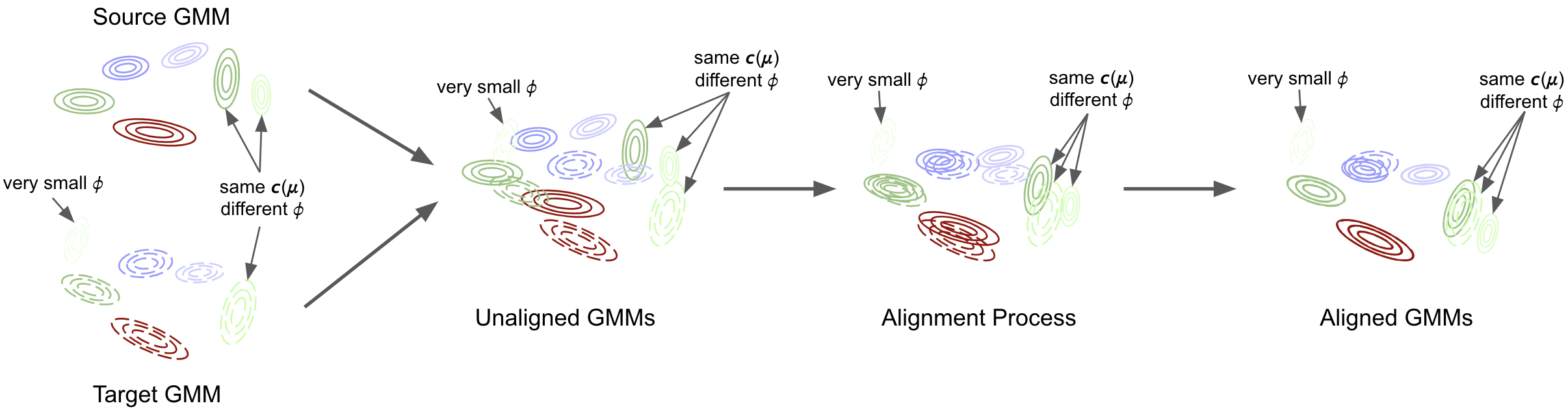}
    \caption{Illustration of the alignment process. The value of the objective decreases as Gaussians align in both position and colour.}
    \label{fig:gmm_align}
\end{figure}

We represent rotation with a unit quaternion $\mathbf{q}$ and translation with a 3D vector $\mathbf{t} \in \mathbb{R}^3$.
The pose is initialised using the estimate provided by MASt3R \cite{MASt3R}, which serves as a strong prior for both rotation and translation.
The objective in \cref{eq:final_obj_conf} is maximised using the Adam optimiser~\cite{Adam}, with separate learning rates $\eta_\text{rot}$ and $\eta_\text{trans}$ for optimising $\mathbf{q}$ and $\mathbf{t}$ respectively, to account for their differing sensitivities and magnitudes.
All key hyperparameters---including $\eta_\text{rot}$, $\eta_\text{trans}$, colour covariance $\bm{\Sigma}^c$, spatial covariance scaling factor $\lambda_\text{cov}$ and others---are determined via Bayesian optimisation on a validation set.
The optimisation procedure is illustrated in \cref{fig:gmm_align}, where it is shown that Gaussians with higher opacity contribute more strongly to the objective.

\section{Experiments}\label{sec:experiments}

We validate our method, \texttt{GARPS}, through extensive experiments, comparing it against classical and state-of-the-art methods on a challenging dataset.

\subsection{Dataset and Metrics}

We use the RealEstate10K dataset \cite{RealEstate10K}, which contains diverse scenes with natural camera motion.
As our method is not a learning-based approach, no training set is used.
We randomly sample 200 clips from the test split:
100 with small viewpoint changes (translation $\leq$ 0.3 m and rotation $\leq$ 15\textdegree) and 100 with large viewpoint changes (translation $>$ 0.3 m or rotation $>$ 15\textdegree), based on camera poses between the 1st and 30th frames.
One image pair is generated per clip for evaluation.
Similarly, we randomly sample 20 clips (10 from each category) from the dataset's training split for hyperparameter tuning.

We evaluate pose accuracy using standard metrics: rotation error ($E_\text{rot}$ in degrees), translation error ($E_\text{trans}$ in metres), and a 3D alignment error ($E_\text{3D}$ in metres), which measures the average displacement of transformed Gaussian centres.
The latter is defined as the average Euclidean distance between the Gaussian centres of the first (source) GMM transformed by the estimated pose, and those transformed by the ground-truth pose, and is given by
\begin{equation}
E_\text{3D} = \frac{1}{N} \sum_{i=1}^{N} \left\| \bm{\hat{R}} \bm{\mu}_i + \bm{\hat{t}} - (\bm{R} \bm{\mu}_i + \bm{t}) \right\|,
\end{equation}
where $\{\bm{\mu}_i\}_{i=1}^N$ is a set of $N$ source Gaussian means,
$(\bm{\hat{R}}, \bm{\hat{t}})$ is the estimated pose, and $(\bm{R}, \bm{t})$ is the ground-truth pose.
This error measures the discrepancy between the estimated and ground-truth transformations by applying both to the same source model and computing the average Euclidean distance between the resulting transformed points.
Since ground-truth point clouds and Gaussian mixtures are unavailable, we use the output of Flash3D as pseudo ground truth geometry for this error measure.

Although Mahalanobis distance may better reflect the GMM structure, we adopt Euclidean distance for $E_\text{3D}$ as it offers a clearer geometric interpretation and avoids further reliance on the predicted covariances from Flash3D.
This choice reflects a trade-off: Mahalanobis distance would fit better with the GMM formulation, but would make the evaluation even more dependent on the quality of the pseudo ground truth.

\subsection{Implementation Details}
Our method is implemented in PyTorch and runs on a single NVIDIA GTX 2080Ti GPU. We used the following key hyperparameters, tuned via Bayesian optimisation on the validation set: 
$\lambda_\text{sem} = 0.92$, 
$\lambda_\text{cov} = 90$, 
$\mathbf{\Sigma}^c = \mathrm{diag}(0.4, 0.4, 0.4)$, 
$\tau_\text{opacity} = 0.5$, 
$\eta_\text{trans} = 9 \times 10^{-5}$, 
$\eta_\text{rot} = 3 \times 10^{-6}$, 
$s_\text{sub} = 32$, 
$k_\text{nn} = 30$, and 
$\tau_\text{dist} = 2.5$.

\subsection{Comparison Methods}
We compare \texttt{GARPS} against classical baselines and recent state-of-the-art methods.
Baselines include the SfM pipeline COLMAP \cite{COLMAP}, geometry-based ICP \cite{ICP} and Colored ICP \cite{ColoredICP}, and the probabilistic CPD \cite{CPD} method.
SOTA methods include DUSt3R \cite{DUSt3R} and MASt3R \cite{MASt3R}, which regress aligned 3D pointmaps.
For point-cloud-based methods, we provide them with the point clouds from Flash3D with DepthAnythingV2 and a pose initialisation using the estimate from MASt3R.

\subsection{Results}\label{sec:results}

We define a successful pair as an image pair for which a method produces a valid relative pose.
All reported statistics are computed only on the successful pairs for each method. 
COLMAP achieves a relatively low success rate of 22.5\%, while all other methods attain a 100\% success rate.

We report both mean $\pm$ standard deviation and median $\pm$ 0.5 $\times$ interquartile range (IQR) of pose estimation errors on 200 samples from the RealEstate10K test set.
\Cref{tab:results_mean,tab:results_median} show that \texttt{GARPS} consistently outperforms both classical and learning-based baselines in most metrics.
While DUSt3R achieves the best median rotation error, \texttt{GARPS} demonstrates robust overall performance and surpasses MASt3R in all metrics.
Unlike the learning-based models, \texttt{GARPS} does not require task-specific training, yet delivers state-of-the-art performance through a purely optimisation-based pipeline.

In terms of runtime, the reported time for \texttt{GARPS} (17.69 seconds) includes the initial pose estimation by MASt3R. 
While \texttt{GARPS} is slower than lightweight geometric methods such as ICP, it is still considerably faster than classical non-rigid methods like CPD.

\begin{figure}[t]
  \centering
  \begin{subfigure}[b]{0.45\textwidth}
    \includegraphics[width=\linewidth]{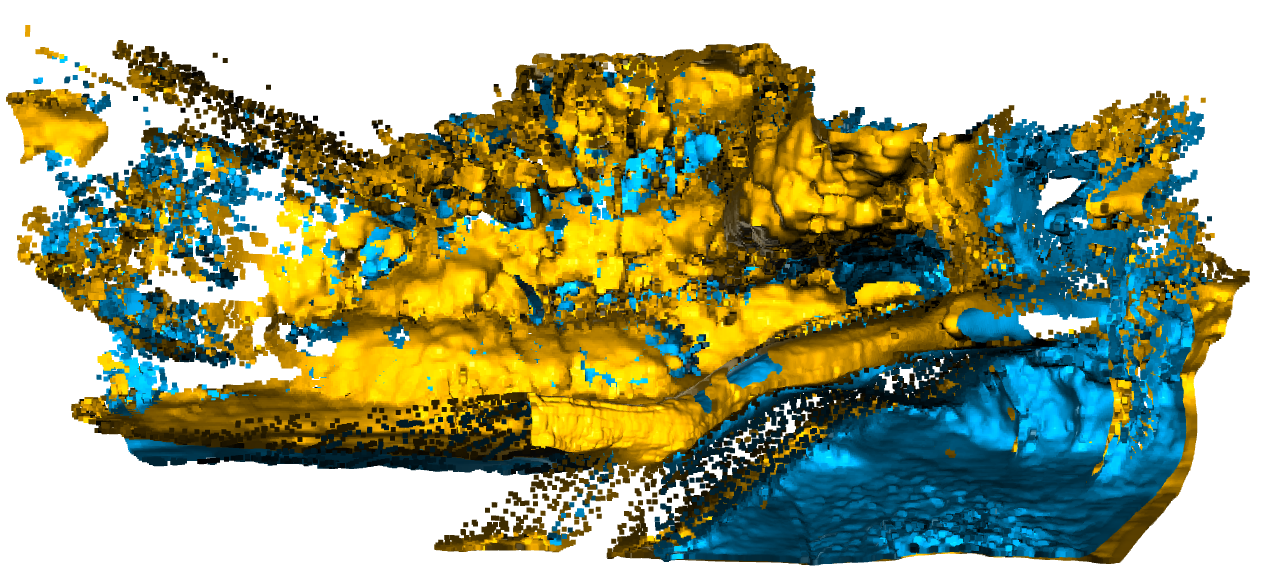}
    \caption{GARPS}
  \end{subfigure}\hfill
  \begin{subfigure}[b]{0.45\textwidth}\includegraphics[width=\linewidth]{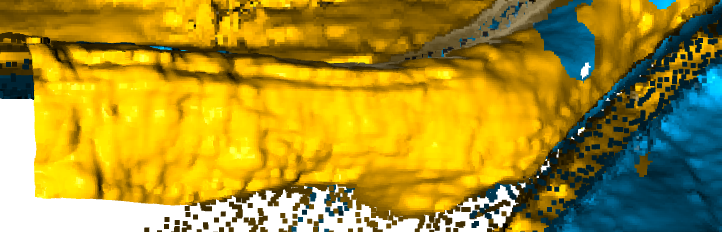}
  \caption{GARPS (Zoomed)}
  \end{subfigure}\vfill
  \begin{subfigure}[b]{0.45\textwidth}
    \includegraphics[width=\linewidth]{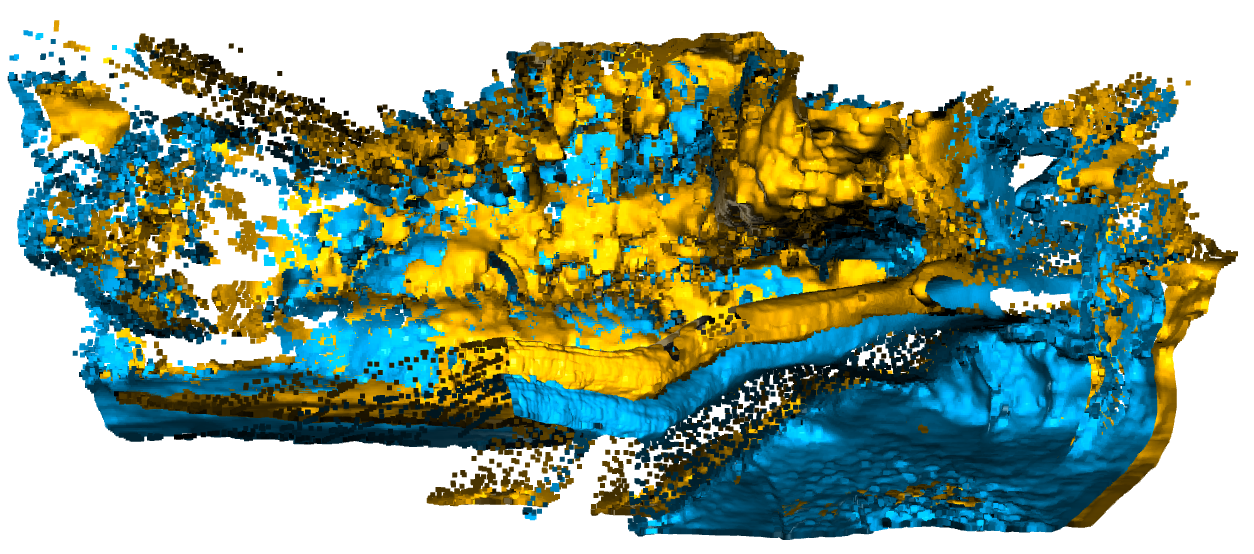}
    \caption{Colored ICP}
  \end{subfigure}\hfill
  \begin{subfigure}[b]{0.45\textwidth}
    \includegraphics[width=\linewidth]{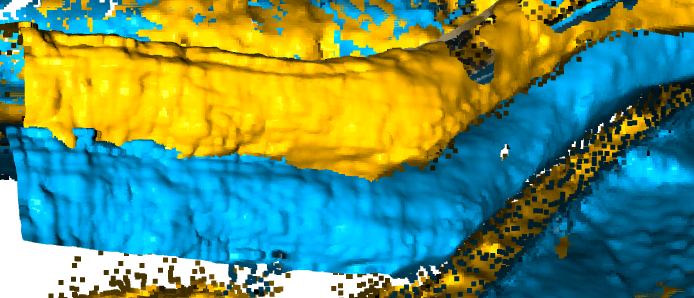}
    \caption{Colored ICP (Zoomed)}
  \end{subfigure}\vfill
  \begin{subfigure}[b]{0.45\textwidth}
    \includegraphics[width=\linewidth]{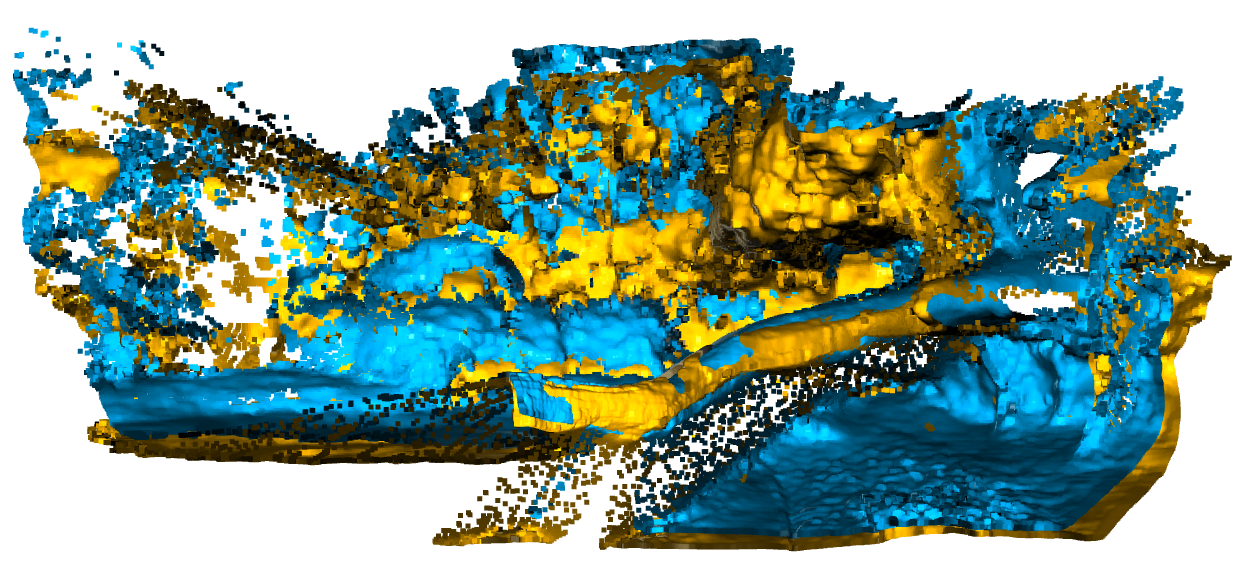}
    \caption{DUSt3R}
  \end{subfigure}\hfill
  \begin{subfigure}[b]{0.45\textwidth}
    \includegraphics[width=\linewidth]{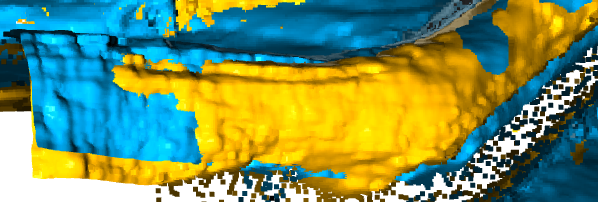}
    \caption{DUSt3R (Zoomed)}
  \end{subfigure}%
  \caption{Qualitative comparison on a test scene (full/zoomed views). \texttt{GARPS} produces better geometric alignment compared to baselines and SOTA methods. The first point cloud is yellow, the second is blue.}
  \label{fig:qualitative_results}
\end{figure}

The qualitative results, shown in \cref{fig:qualitative_results}, support the quantitative findings.
\texttt{GARPS} produces sharper and more consistent geometric alignment---for example, in challenging regions such as trees and trunks---while other methods exhibit visible translational or rotational drift.

\begin{table}[t]
  \centering
  \caption{Mean pose errors (± standard deviation) and average inference time per pair on the test set.}
  \label{tab:results_mean}
  \begin{tabularx}{\linewidth}{@{}lCCCc@{}}
    \toprule
    Method & $E_\text{trans} \downarrow$ (m) & $E_\text{rot} \downarrow$ (deg) & $E_\text{3D} \downarrow$ (m) & Time (s)\\
    \midrule
    COLMAP      & 0.251 ± 0.274 & 9.422 ± 11.037 & 0.492 ± 0.451 & 3.18 \\
    ICP         & 0.482 ± 0.628 & 6.246 ± 14.470 & 0.211 ± 0.257 & \textbf{0.04} \\
    Colored ICP & 0.427 ± 0.506 & 5.693 ± 12.869 & 0.189 ± 0.205 & 0.06 \\
    CPD         & 0.492 ± 0.616 & 9.981 ± 10.328 & 0.506 ± 0.471 & 266 \\
    \midrule
    DUSt3R      & 0.345 ± 0.278 & 1.058 ± 1.628 & 0.171 ± 0.133 & 13.07 \\
    MASt3R      & 0.170 ± 0.200 & 0.995 ± 1.549 & 0.097 ± 0.110 & 9.07 \\
    \midrule
    \texttt{GARPS} & \textbf{0.167 ± 0.200} & \textbf{0.994 ± 1.545} & \textbf{0.094 ± 0.109} & 17.69 \\
    \bottomrule
  \end{tabularx}
\end{table}

\begin{table}[t]
  \centering
  \caption{Median pose errors ($\pm$ 0.5 IQR) on the test set.}
  \label{tab:results_median}
  \begin{tabularx}{\linewidth}{@{}lCCC@{}}
    \toprule
    Method & $E_\text{trans} \downarrow$ (m) & $E_\text{rot} \downarrow$ (deg) & $E_\text{3D} \downarrow$ (m) \\
    \midrule
    COLMAP      & 0.166 ± 0.104 & 5.207 ± 7.510 & 0.386 ± 0.251 \\
    ICP         & 0.282 ± 0.202 & 2.607 ± 1.994 & 0.127 ± 0.086 \\
    Colored ICP & 0.275 ± 0.188 & 2.434 ± 1.822 & 0.125 ± 0.083 \\
    CPD         & 0.316 ± 0.180 & 7.406 ± 5.712 & 0.397 ± 0.255 \\
    \midrule
    DUSt3R      & 0.262 ± 0.152 & \textbf{0.522 ± 0.371} & 0.140 ± 0.072 \\
    MASt3R      & 0.114 ± 0.066 & 0.570 ± 0.421 & 0.062 ± 0.031 \\
    \midrule
    \texttt{GARPS} & \textbf{0.109 ± 0.065} & 0.561 ± 0.400 & \textbf{0.058 ± 0.030} \\
    \bottomrule
  \end{tabularx}
\end{table}

\subsection{Ablation Study}

\begin{table}[t]
  \centering
  \caption{Ablation study on the test set (mean pose errors). All values indicate relative change (\%) with respect to the full method.}
  \label{tab:ablation_results}
  \begin{tabularx}{\linewidth}{@{}lCCC@{}}
    \toprule
    Variant & $E_\text{trans} \downarrow$ & $E_\text{rot} \downarrow$ & $E_\text{3D} \downarrow$ \\
    \midrule
    w/o image segmentation & +0.05\% & +0.10\% & +0.02\% \\
    w/o anisotropic covariances & +1.01\% & +0.71\% & +2.41\%  \\
    w/o colour features & +0.45\% & +0.58\% & +1.11\% \\
    \bottomrule
  \end{tabularx}
\end{table}

The ablation study in \cref{tab:ablation_results} highlights the contribution of each major component in our framework.
Each variant removes one key element from the full method to isolate its impact on the performance, measured by the relative change (\%) in mean pose errors: translation error ($E_\text{trans}$), rotation error ($E_\text{rot}$), and 3D error ($E_\text{3D}$).
The omission of the image segmentation leads to a minor degradation, indicating that it complements other modules without being the sole performance driver. 
Replacing the original anisotropic covariance in Flash3D with an isotropic covariance (i.e., an identity matrix) results in a significant performance drop, suggesting that modelling uncertainty plays a critical role in accurate pose estimation.
Similarly, omitting colour information causes noticeable degradation across all metrics, highlighting its complementary role in enhancing geometric understanding.
These results validate the effectiveness and necessity of each proposed component.

\section{Conclusion}\label{sec:conclusion}

We present GARPS, a novel method for estimating metrically-scaled relative pose from two RGB images by aligning single-view 3D Gaussian mixtures.
Our method leverages monocular 3D reconstructions from Flash3D~\cite{Flash3D}, enhanced with Depth Anything V2~\cite{DepthAnythingV2}, semantic labels from DINOv2~\cite{DINOv2}, and prior pose estimates from MASt3R~\cite{MASt3R}.
It aligns two independently generated coloured 3D GMMs using a robust $L_2$ objective that jointly considers geometry, colour, semantics, and anisotropic spatial uncertainties.
\texttt{GARPS} sets a new state-of-the-art in relative pose estimation from two-view RGB images, outperforming both classical and learning-based methods across all evaluated metrics.

Future work includes exploring monocular reconstruction methods beyond Flash3D that yield 3D GMMs, simplifying covariance computations to reduce runtime, and extending the method to dynamic scenes via object tracking.


%
%
%
\bibliographystyle{splncs04}
\bibliography{bib}

\begin{thebibliography}{10}
\providecommand{\url}[1]{\texttt{#1}}
\providecommand{\urlprefix}{URL }
\providecommand{\doi}[1]{https://doi.org/#1}

\bibitem{Texture-Gradient}
Bajcsy, R., Lieberman, L.: Texture gradient as a depth cue. Computer Graphics and Image Processing  \textbf{5}(1),  52--67 (1976)

\bibitem{ICP}
Besl, P.J., McKay, N.D.: A method for registration of 3-d shapes. IEEE Trans. Pattern Anal. Mach. Intell.  \textbf{14}(2),  239–256 (Feb 1992)

\bibitem{GOGMA}
Campbell, D., Petersson, L.: Gogma: Globally-optimal gaussian mixture alignment. In: 2016 IEEE Conference on Computer Vision and Pattern Recognition (CVPR). pp. 5685--5694 (2016). \doi{10.1109/CVPR.2016.613}

\bibitem{SuperPoint}
DeTone, D., Malisiewicz, T., Rabinovich, A.: Superpoint: Self-supervised interest point detection and description. In: Proceedings of the IEEE conference on computer vision and pattern recognition workshops. pp. 224--236 (2018)

\bibitem{RANSAC}
Fischler, M.A., Bolles, R.C.: Random sample consensus: A paradigm for model fitting with applications to image analysis and automated cartography. In: Fischler, M.A., Firschein, O. (eds.) Readings in Computer Vision, pp. 726--740. Morgan Kaufmann, San Francisco (CA) (1987)

\bibitem{hartley2003multiple}
Hartley, R.: Multiple view geometry in computer vision, vol.~665. Cambridge university press (2003)

\bibitem{GMMReg}
Jian, B., Vemuri, B.C.: Robust point set registration using gaussian mixture models. IEEE transactions on pattern analysis and machine intelligence  \textbf{33}(8),  1633--1645 (2010)

\bibitem{3DGS}
Kerbl, B., Kopanas, G., Leimk{\"u}hler, T., Drettakis, G.: 3d gaussian splatting for real-time radiance field rendering. ACM Trans. Graph.  \textbf{42}(4),  139--1 (2023)

\bibitem{Adam}
Kingma, D.P., Ba, J.: Adam: A method for stochastic optimization (2017), \url{https://arxiv.org/abs/1412.6980}

\bibitem{MASt3R}
Leroy, V., Cabon, Y., Revaud, J.: Grounding image matching in 3d with mast3r (2024)

\bibitem{LightGlue}
Lindenberger, P., Sarlin, P.E., Pollefeys, M.: {LightGlue: Local Feature Matching at Light Speed}. In: ICCV (2023)

\bibitem{GS-CPR}
Liu, C., Chen, S., Bhalgat, Y.S., HU, S., Cheng, M., Wang, Z., Prisacariu, V.A., Braud, T.: {GS}-{CPR}: Efficient camera pose refinement via 3d gaussian splatting. In: The Thirteenth International Conference on Learning Representations (2025), \url{https://openreview.net/forum?id=mP7uV59iJM}

\bibitem{SIFT}
Lowe, D.G.: Distinctive image features from scale-invariant keypoints. International journal of computer vision  \textbf{60},  91--110 (2004)

\bibitem{NeRF}
Mildenhall, B., Srinivasan, P.P., Tancik, M., Barron, J.T., Ramamoorthi, R., Ng, R.: Nerf: Representing scenes as neural radiance fields for view synthesis. In: ECCV (2020)

\bibitem{CPD}
Myronenko, A., Song, X., Carreira-Perpinan, M.: Non-rigid point set registration: Coherent point drift. Advances in neural information processing systems  \textbf{19} (2006)

\bibitem{Nister2004}
Nister, D.: An efficient solution to the five-point relative pose problem. IEEE Transactions on Pattern Analysis and Machine Intelligence  \textbf{26}(6),  756--770 (2004)

\bibitem{DINOv2}
Oquab, M., Darcet, T., Moutakanni, T., Vo, H.V., Szafraniec, M., Khalidov, V., Fernandez, P., Haziza, D., Massa, F., El-Nouby, A., Howes, R., Huang, P.Y., Xu, H., Sharma, V., Li, S.W., Galuba, W., Rabbat, M., Assran, M., Ballas, N., Synnaeve, G., Misra, I., Jegou, H., Mairal, J., Labatut, P., Joulin, A., Bojanowski, P.: Dinov2: Learning robust visual features without supervision (2023)

\bibitem{ColoredICP}
Park, J., Zhou, Q.Y., Koltun, V.: Colored point cloud registration revisited. In: 2017 IEEE International Conference on Computer Vision (ICCV). pp. 143--152 (2017). \doi{10.1109/ICCV.2017.25}

\bibitem{SuperGlue}
Sarlin, P.E., DeTone, D., Malisiewicz, T., Rabinovich, A.: Superglue: Learning feature matching with graph neural networks. In: Proceedings of the IEEE/CVF conference on computer vision and pattern recognition. pp. 4938--4947 (2020)

\bibitem{COLMAP}
Sch\"{o}nberger, J.L., Frahm, J.M.: Structure-from-motion revisited. In: Conference on Computer Vision and Pattern Recognition (CVPR) (2016)

\bibitem{COLMAP_MVS}
Sch\"{o}nberger, J.L., Zheng, E., Pollefeys, M., Frahm, J.M.: Pixelwise view selection for unstructured multi-view stereo. In: European Conference on Computer Vision (ECCV) (2016)

\bibitem{szeliski2022computer}
Szeliski, R.: Computer vision: algorithms and applications. Springer Nature (2022)

\bibitem{Flash3D}
Szymanowicz, S., Insafutdinov, E., Zheng, C., Campbell, D., Henriques, J.F., Rupprecht, C., Vedaldi, A.: Flash3d: Feed-forward generalisable 3d scene reconstruction from a single image (2024), \url{https://arxiv.org/abs/2406.04343}

\bibitem{SplatterImage}
Szymanowicz, S., Rupprecht, C., Vedaldi, A.: Splatter image: Ultra-fast single-view 3d reconstruction. In: The IEEE/CVF Conference on Computer Vision and Pattern Recognition (CVPR) (2024)

\bibitem{DUSt3R}
Wang, S., Leroy, V., Cabon, Y., Chidlovskii, B., Revaud, J.: Dust3r: Geometric 3d vision made easy. In: Proceedings of the IEEE/CVF Conference on Computer Vision and Pattern Recognition (CVPR). pp. 20697--20709 (June 2024)

\bibitem{Go-ICP}
Yang, J., Li, H., Campbell, D., Jia, Y.: Go-icp: A globally optimal solution to 3d icp point-set registration. IEEE Transactions on Pattern Analysis and Machine Intelligence  \textbf{38}(11),  2241--2254 (2016). \doi{10.1109/TPAMI.2015.2513405}

\bibitem{DepthAnythingV2}
Yang, L., Kang, B., Huang, Z., Zhao, Z., Xu, X., Feng, J., Zhao, H.: Depth anything v2 (2024), \url{https://arxiv.org/abs/2406.09414}

\bibitem{Shape-from-Shading}
Zhang, R., Tsai, P.S., Cryer, J., Shah, M.: Shape-from-shading: a survey. IEEE Transactions on Pattern Analysis and Machine Intelligence  \textbf{21}(8),  690--706 (1999)

\bibitem{Flare}
Zhang, S., Wang, J., Xu, Y., Xue, N., Rupprecht, C., Zhou, X., Shen, Y., Wetzstein, G.: Flare: Feed-forward geometry, appearance and camera estimation from uncalibrated sparse views. arXiv preprint arXiv:2502.12138  (2025)

\bibitem{ADE20K_2}
Zhou, B., Zhao, H., Puig, X., Fidler, S., Barriuso, A., Torralba, A.: Scene parsing through ade20k dataset. In: Proceedings of the IEEE Conference on Computer Vision and Pattern Recognition (2017)

\bibitem{ADE20K}
Zhou, B., Zhao, H., Puig, X., Xiao, T., Fidler, S., Barriuso, A., Torralba, A.: Semantic understanding of scenes through the ade20k dataset. International Journal of Computer Vision  \textbf{127}(3),  302--321 (2019)

\bibitem{RealEstate10K}
Zhou, T., Tucker, R., Flynn, J., Fyffe, G., Snavely, N.: Stereo magnification: Learning view synthesis using multiplane images. In: SIGGRAPH (2018)

\bibitem{LoopSplat}
Zhu, L., Li, Y., Sandström, E., Huang, S., Schindler, K., Armeni, I.: Loopsplat: Loop closure by registering 3d gaussian splats. In: International Conference on 3D Vision (3DV) (2025)

\end{thebibliography}
%




\end{document}